\documentclass[11pt,onecolumn]{article}

\usepackage{booktabs}       
\usepackage{caption} 
\usepackage{footnote}
\makesavenoteenv{tabular}
\usepackage[utf8]{inputenc}
\usepackage{times}
\usepackage{amsmath}
\usepackage{amssymb}
\usepackage{multirow}
\usepackage{xspace}
\usepackage{theorem}
\usepackage{graphicx}
\usepackage{ifpdf}
\usepackage{url,hyperref}
\usepackage{latexsym}
\usepackage{euscript}
\usepackage{xspace}
\usepackage{color}
\usepackage{makeidx}
\usepackage{wrapfig}
\usepackage{xypic}
\xyoption{all}

\long\def\remove#1{}

\newif\ifverbose
\newif\ifcomments

\usepackage{subfiles} 
\usepackage{graphicx} 
\usepackage{amsmath}
\usepackage{amsfonts}
\newcommand{\citep}{\cite}
\newcommand{\citet}{\cite}
\usepackage{subcaption}
\usepackage{graphicx}
\usepackage{wrapfig}
\newcommand{\PD}{PD }
\newcommand{\PDs}{PDs }

\title{Understanding the Power of Persistence Pairing via Permutation Test}
\author{Chen Cai \\
The Ohio State University\\
\texttt{cai.507@osu.edu} \\
\and
Yusu Wang  \\
The Ohio State University\\
\texttt{yusu@cse.ohio-state.edu} \\
}
\date{}

\begin{document}

\maketitle

\begin{abstract}
  Recently many efforts have been made to incorporate persistence diagrams, one of major tools in topological data analysis (TDA), into machine learning pipelines. 
  To better understand the power and limitation of persistence diagrams, we carry out a range of experiments on both graph data and shape data, aiming to decouple and inspect the effects of different factors involved. To this end, we also propose the so-called \emph{permutation test} for persistence diagrams to delineate critical values and pairings of critical values. 
  For graph classification tasks, we note that while persistence pairing yields consistent improvement over various benchmark datasets, it appears that for various filtration functions tested, most discriminative power comes from critical values. For shape segmentation and classification, however, we note that persistence pairing shows significant power on most of the benchmark datasets, and improves over both summaries based on merely critical values, and those based on permutation tests.  Our results help provide insights on when persistence diagram based summaries could be more suitable. 
  \end{abstract}
\section{Introduction}
Topological data analysis (TDA) is an emerging field that aims to characterize the shape of low and high dimensional data via methods steming from algebraic topology. One of the major tools of TDA is persistence diagrams (PDs). Given a function on the manifold, \PD can concisely summarize the birth and death of topological features (connected components, loops, cavities...) of (sub-/super-) level sets with respect to the function. \PDs also enjoy the stability property \cite{cohen2007stability} that makes it potentially useful for machine learning. 


In recent years, many efforts have been made to utilize \PDs as features for downstream machine learning tasks, such as material science \cite{buchet2018persistent}, signal analysis \cite{perea2015sliding} , cellular data \cite{camara2017topological} and shape recognition \cite{li2014persistence}. However, the geometry of the \PD does not lend itself easily to well-adopted classifiers due to the lack of Hilbert structure. In particular, several basic operations, such as mean and addition, are not well defined \cite{turner2014frechet} for PD, making it difficult to utilize \PDs straightforwardly in machine learning. 

To handle this issue, a natural way is to apply vectorization \cite{bubenik2015statistical, carriere2015stable, chazal2014stochastic, kalivsnik2019tropical} or kernelization \cite{reininghaus2015stable, carriere2017sliced} to PDs, i.e., embedding \PDs either to a Euclidean space $\mathbb{R}^d$ or a reproducing kernel Hilbert space (RKHS) associated with certain kernels. However, these approaches still have limitations. First, it has been shown that finite-dimension embedding can miss information about \PDs \cite{carriere2018metric}. Second, the time of computing a kernel is quadratic in the number of PDs, which is quite expensive for large scale applications. Third, choosing right vectorization/kernelization and its associated hyper-parameters is not straightforward and usually requires multiple rounds of trial and error. 

Due to these extra complexities, one may wonder whether the benefits of using \PDs outweigh the extra cost. Specifically, \PDs have two major components: 1) filtration function and 2) persistence pairing (decomposition of persistence module). 
In this paper, we ask a simple yet fundamental question: \emph{How much extra power can persistence pairing  bring in?} 

 A straightforward way is to look at results obtained from statistics of filtration function (e.g., histogram) versus the PDs. However, due to the format mismatch between vectors and PDs, directly comparing them may be affected by other design choices such as what kernel (or distance measures) to use and associated hyper-parameters. To better measure the power of PDs, we propose a simple trick named ``permutation test'' that interpolate histograms of filtration function and PDs. 

  Specifically, we permute the persistence points in such a way that only coordinates of \PDs remain the same but the original pairing is completely destroyed. These fake PDs have the same form as the original PDs and therefore the same kernels for true diagrams can also be applied to fake ones. We use these fake diagrams as the input for various tasks and check their effectiveness. As we will see, this simple trick brings various insights on the use of \PDs for different problems. 

\begin{figure}[h]
\begin{center} 
\begin{tabular}{cccc}
\includegraphics[height=4cm]{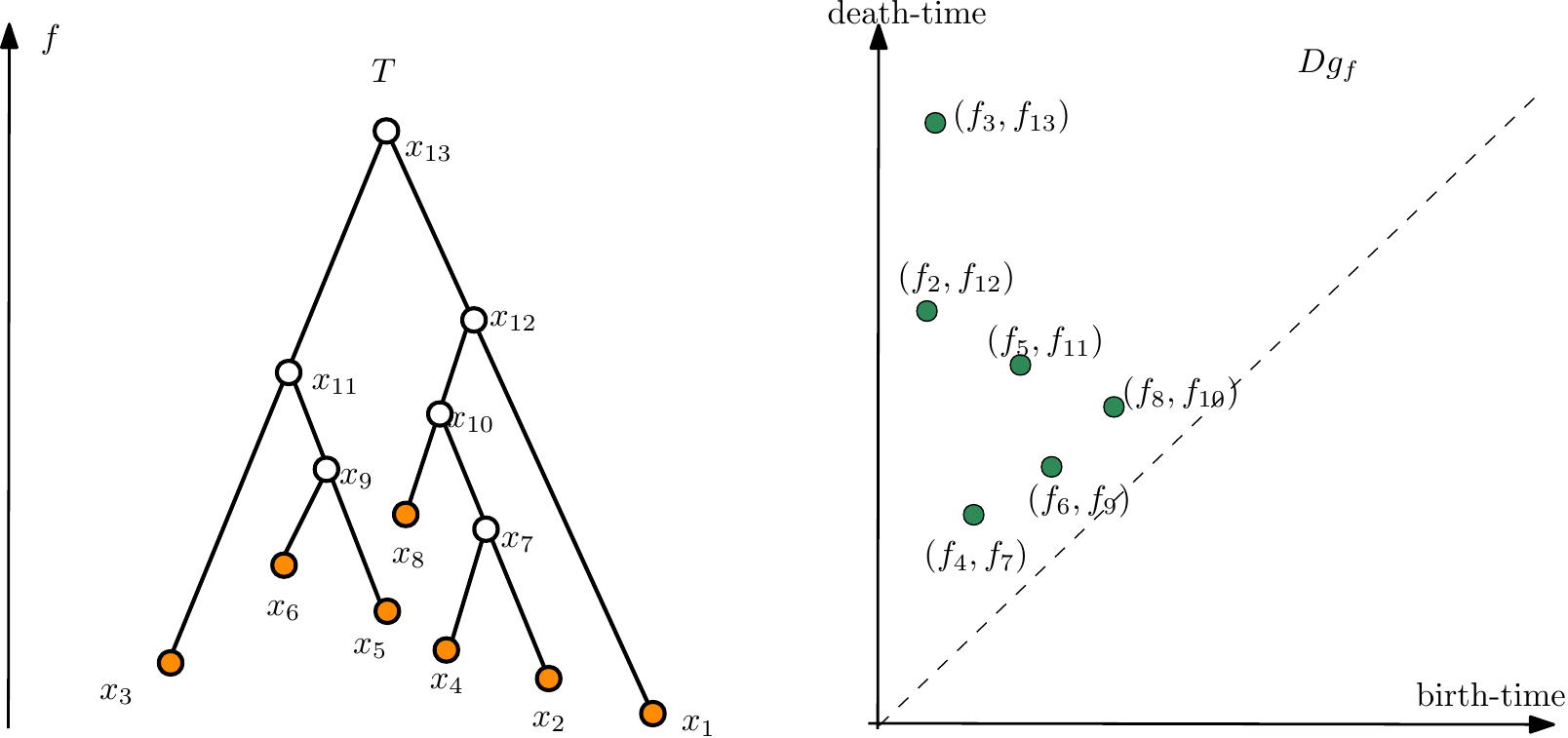} &
\end{tabular}
\end{center}
\caption{Given the function $f$ as the height, the induced sublevel \PD is shown on the right side. 
}

\end{figure}
\textbf{Our Contributions.} To the best of our knowledge, our paper is the first work systematically quantifying (empirically) the power of persistence pairing for various applications. Specifically, our contributions are the following.
\begin{itemize}

\item We propose the permutation test for \PDs that decouples the statistics of the critical values of filtration function and the persistence pairing. Using the proposed permutation test, we find that in graph classification, even fake diagrams perform quite well compared to original \PDs. We believe this is due to the noisy nature of graph datasets (\PDs are not stable against random insertion and deletion of edges).

As a byproduct of our extensive experiments, we also provide some rules of thumb for using \PDs in graph classification.

\item For shape segmentation and classification, we find the power of persistence pairing depends on the particular featurization chosen. With the right choice of featurization, utilizing persistence pairing brings in significant improvement. Intuitively, we think that the shape models have more prominent geometric features in them, which are effectively captured by PDs. In contrast, PDs seem to be less effective at capturing features for graphs, partly due to the choice of descriptor functions as well as the nature of noise in graph (e.g., random insertions) which makes \PDs less stable. 

\item We study the structure of true diagrams and fake diagrams via confusion matrix analysis and visualization. Fake diagrams are not only shown to be well separated from true diagrams but also seem to separate different classes reasonably well for graph classification.

\end{itemize}

\section{Background}
\subsection{Persistent Homology}
The definition of our proposed method relies on the so-called persistence diagram induced by a scalar function. We refer readers to resources such as \cite{edelsbrunner2010computational, oudot2015persistence} for formal discussions on persistent homology and related developments. Below we only provide an intuitive and informal description of the persistent homology induced by a function under a simple setting. Let $f : X \rightarrow \mathbb{R}$ be a continuous real-valued function defined on a topological space $X$. We want to understand the structure of $X$ from the perspective of $f$. Specifically, let $X_{\alpha}:= \{x \in X | f(x) < \alpha \}$ denote the sublevel set of $X$ w.r.t. $\alpha \in \mathbb{R}$. Now as we sweep $X$ bottom-up (top down) by increasing the value, the sequence of sublevel (superlevel) sets connected by natural inclusion maps gives rise to a filtration of $X$ induced by $f$:
\begin{equation}\label{eqn:sublevelseq}
X_{\alpha_{1}} \subset X_{\alpha_{2}}  \subset ... \subset X_{\alpha_{m}} =X, \alpha_{1} < \alpha_{2} < ... < \alpha_{m}
\end{equation}
We track how the topological features of sublevel sets change in terms of homology classes. In particular, as $\alpha$ increases, sometimes new topological features are born at time $\alpha$, that is, new families of homology classes are created in $H_{k}(X_{\alpha})$, the $k$-th homology group of $X$. Sometimes, existing topological features disappear, i.e, some homology classes become trivial in $H_{k}(X_{\beta})$ for some $ \beta > \alpha$. The persistent homology captures such birth and death events, and summarizes them in the so-called persistence diagram $Dg_{k}(f)$ ( We will call it diagram for short when no ambiguity is raised). Specifically, $Dg_{k}(f)$ consists of a set of persistence points $(\alpha, \beta) \in \mathbb{R}^2$, where each $(\alpha, \beta)$ indicates a $k$-th homological feature created at $\alpha$ and killed at $\beta$. we also call $(\alpha, \beta)$ persistence pairing since it pairs two critical values $\alpha$ and $\beta$ of the filtration function.

In particular, 0 homology is just connected components and can be computed efficiently in $O(\alpha(n) n)$ using union-find data structure where $\alpha(n)$ is an extremely slow-growing inverse Ackermann function. 

\subsection{Extended Persistence Homology}
In some context, ordinary persistence may be insufficient to encode the topology of an object $X$. For example, when $X$ is a graph, the loops persist forever since they are not filled during the sublevel filtration. Similarly, upfork branching points (w.r.t. the filtration function $f$) are not captured (while those pointing downwards are detected), since they do not create connected components when they appear in the sublevel filtration.

To address the issues above, extended persistence refines the analysis by also including the super-level set $X^{\alpha} = \{x \in X: f(x) \geq \alpha \}$ into the filtration in Eqn (\ref{eqn:sublevelseq}). Similarly, letting $\alpha$ decrease from $\infty$ to $-\infty$ gives a sequence of increasing subsets, for which structural changes can be recorded. 
In particular, assuming we have a sequence of reals  $\alpha_1 < \alpha_2 < \cdots < \alpha_m$ such that $X_{\alpha_1}=\emptyset$ ($X^{\alpha_1} = X$) and $X_{\alpha_m} = X$ ($X^{\alpha_m} = \emptyset$), we consider the following extended sequence: 
\begin{align}
\emptyset = & X_{\alpha_1} \subseteq X_{\alpha_2} \subseteq \cdots \subseteq X_{\alpha_m} = (X; X^{\alpha_m}) \label{eqn:extendedseq} \\
\subseteq & (X; X^{\alpha_{m-1}}) \subseteq \cdots \subseteq (X; X^{\alpha_2}) \subseteq (X; X^{\alpha_1}) = \emptyset. \nonumber
\end{align}

where $(A; B)$ denotes a pair of space, and at the homology level, note that the natural map from $ X_{\alpha_m} \to (X; X^{\alpha_m})$ induces an isomorphism. 
One can then consider the resulting persistent homology induced by the above extended sequence, and the resulting \PDs are called extended PDs. 
Persistence pairings in such diagrams have four types, depending on whether the birth and death happen during the upward filtration (first line in Eqn (\ref{eqn:extendedseq})) or downward filtration (second line in Eqn (\ref{eqn:extendedseq})). 
In the context of graphs, these types are denoted as $Ord_0$, $Rel_1$, $Ext_{0}^{+}$
and $Ext_{1}^{-}$ for downwards branches, upwards branches, connected components and loops respectively.
Overall, we denote $Dg(G, f) = Ord_0(G, f) \cup Rel_1(G,f) \cup Ext_0^+(G,f) \cup Ext_1^-(G,f).$

\begin{figure}[h]
\begin{center}
\begin{tabular}{cccc}
\includegraphics[height=4cm]{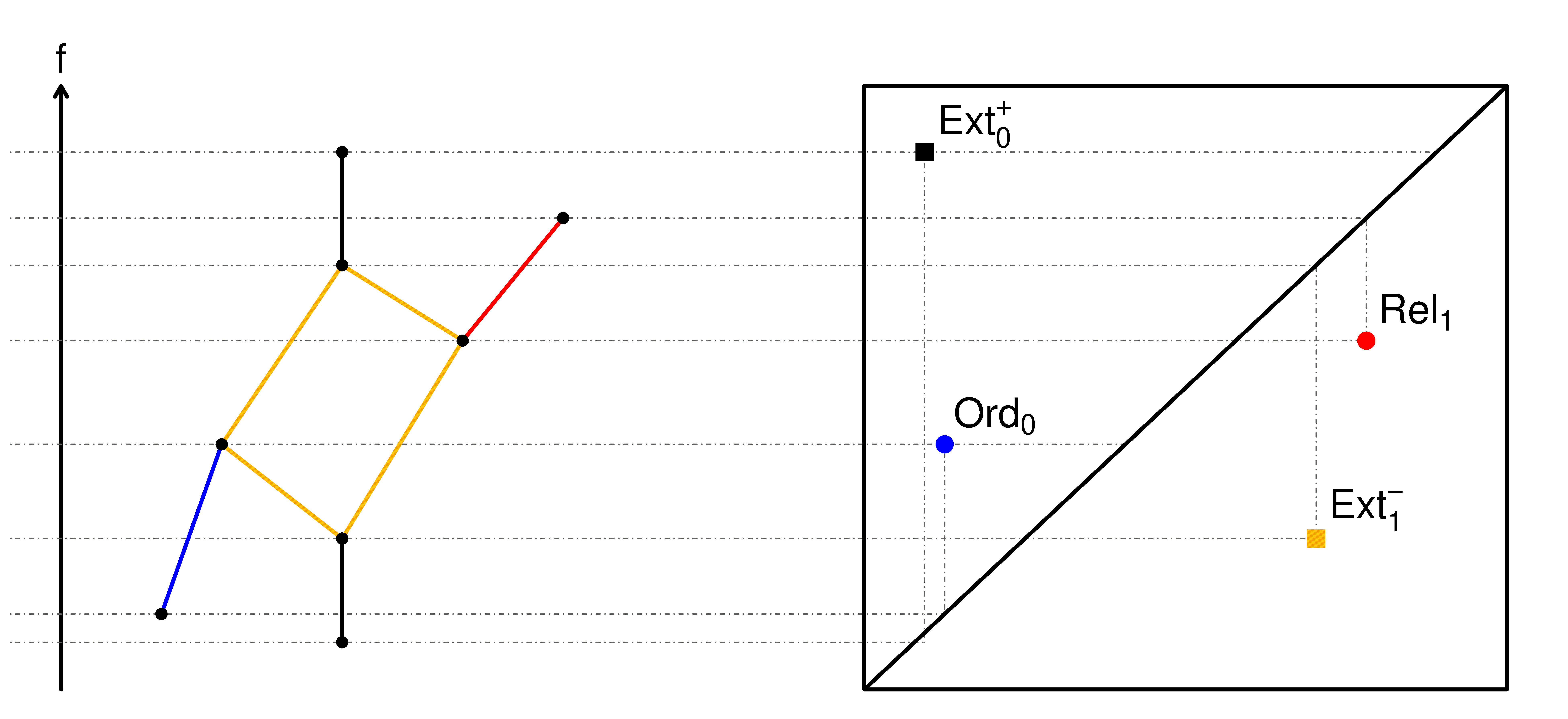} &
\end{tabular}
\end{center}
\caption{Given the filtration $f$ as height for graph $X$, the four types of features of graphs and their corresponding persistence points in the extended persistence diagram.}
\end{figure}

\subsection{\PDs for Machine Learning}
\PDs have been proposed as features for machine learning in a series of work (e.g, \cite{adams2017persistence, chazal2014stochastic}), starting from persistence landscapes \cite{bubenik2015statistical}. We briefly review related methods in this section. The detailed information can be found in the appendix.

\textbf{Kernel Method}. Given a set $\mathcal{X}$ (PDs in our case, a function $k : \mathcal{X}\times \mathcal{X} \rightarrow \mathbb{R}$ is called a positive definite kernel if for all integers $n$ and all families $x_1, ..., x_n$ of $n$ elements in $X$, the matrix $[k(x_i; x_j )]_{i, j}$  itself is positive semi-definite. 
It is known that kernels generalize scalar products, in the sense that, given a kernel $k$, there exists a RKHS $\mathcal{H}_k$ and a feature map $\phi: \mathcal{X} \rightarrow \mathcal{H}_k$ such that $k(x_1, x_2) = <\phi(x_1), \phi(x_2)>_{\mathcal{H}_k}$. A kernel $k$ also induces a distance $d_k$ that can be computed as the Hilbert norm of the difference between two embeddings: $d_k^2(x_1, x_2) = k(x_1, x_1) + k(x_2, x_2) - 2k(x_1, x_2)$. 

In this paper, we try four common kernels for diagrams, i.e.  Persistence Weighted Gaussian Kernel (pwg) \cite{kusano2016persistence}, Persistence Scale Space Kernel (pss) \cite{reininghaus2015stable}, Sliced Wasserstein Kernel (sw) \cite{carriere2017sliced} and Persistence Fisher Kernel (pf) \cite{le2018persistence}. 

\textbf{Vector Method}.
Another way to use PDs for machine learning is to convert them into vectors. Persistence Landscape (PL)  and Persistence Image (PI)  are two examples. One advantage of vectorization over kernelization is that computing kernels takes quadratic time in the number of diagrams while vectorizing \PDs takes only linear time. Thus, we only use vector methods for shape segmentation where the number of diagrams is roughly 20k-50k.

\subsection{\PDs for Graphs}
For graph classification, \cite{hofer2017deep} is the first work introducing a neural network framework to convert \PDs into feature vectors for graph classification in an end-to-end data-dependent way. Perslay \cite{Mathieu2019Perslay} unifies many existing featurization such as persistence landscape, persistence silhouette \cite{chazal2014stochastic} and persistence surface as different instances of a single permutation invariant neural network based on DeepSets \cite{zaheer2017deep}. Weighted Persistence Image Kernel \cite{zhao2019learning} (WKPI) is a recently proposed weighted kernel based on persistence image. WKPI assigns weights for different locations in \PDs where weights are learned from data via gradient descent. Empirically, improved performance over other kernels is shown for the task of graph classification.

\section{Experiment}
\subsection{Setup}
On a high level, we use \PDs obtained from different filtration functions as features. Depending on the task, we choose a proper featurization method (we use kernel methods when possible since they tend to perform better, but for large scale applications, computing kernels is not feasible so we use vector methods), followed by SVM for final classification. We maintain the same experiment settings for original \PDs and permuted ones. For a comprehensive evaluation, we perform experiments on various tasks (graph classification, shape segmentation, and object classification) and diverse datatypes including social networks, molecules/proteins, and shapes of different categories.

We test degree, Ricci Curvature \cite{lin2011ricci}, closeness centrality and square of Fiedler vector (the eigenvector corresponding to the second smallest eigenvalue of graph Laplacian) for graph classification. On shape datasets, we use geodesic distance as filtration function for shape segmentation and closeness centrality for shape classification (Ricci curvature and heat kernel signature \cite{sun2009concise} are also tested for shape classification but their performances are rather poor). 

We use Dionysus\footnote{http://mrzv.org/software/dionysus2/} to compute \PDs and sklean\_tda \footnote{https://github.com/MathieuCarriere/sklearn\_tda} for kernel computation. Ricci curvature is computed via the code in \cite{ni2015ricci}. Our code is available on Github.\footnote{https://github.com/Chen-Cai-OSU/Esme}.


\subsection{Permutation Test and Baselines}
In this section, we introduce permutation test that aims to preserve statistics of filtration function but destroy the persistence pairing of critical values of the filtration function. For a diagram $P$ of $n$ persistence points $P = \{p_{1}, p_{2}, ..., p_{n}\}$, denote the coordinates of point $p_{i}$ by $(x_{i}, y_{i})$. Take the multiset $P_{\textit{multiset}} = \{x_{1}, y_{1}, x_{2}, y_{2},..., x_{n}, y_{n} \}$ as input. We then randomly sample two values without replacement from $P_\textit{multiset}$ as the coordinates of the persistence point in the fake diagram. Repeat the same procedure $n$ times and get a fake diagram called $P_{fake}$ of size $n$.


To better quantify the power of persistence pairing, we also introduce two baselines. $\mathbf{Pervec}$ (vector obtained from coordinates of points in the $\mathbf{per}$muted diagram) is a histogram vector of the coordinates of PD, i.e., the histogram of $P$. $\mathbf{Filvec}$ is a histogram vector of all the $\mathbf{fil}$tration function values on the graph. The length of Pervec and Filvec is a hyper-parameter chosen from $\{100, 200, 300\}$ by cross-validation.

Intuitively, $\mathbf{Filvec}$ can be thought of a summary purely based on values of the descriptor function. In contrast, the fake \PDs still maintain the form as a \PDs (namely, each diagram is still a multiset of points), while having the same distribution of critical values as the true diagrams. In other words, the fake diagrams obtained from permutation test destroy the specific pairing patterns in \PDs, yet still having the form of PDs: the latter means that we could use the same kernel/distance for original diagrams to compare the fake PDs, thereby removing this factor when we compare the effect of \PDs in different tasks. 
\ifcomments
\chen{polish}
\fi 

\begin{figure}[h]
\begin{center}
\begin{tabular}{cccc}
\includegraphics[height=4cm]{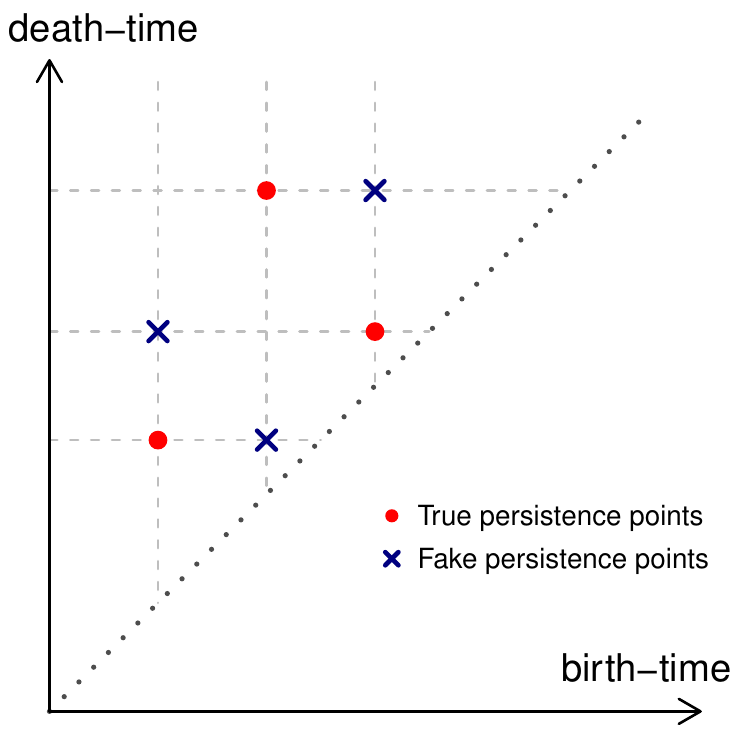} &
\end{tabular}
\end{center}
\caption{Visualize permutation test. The true and fake persistence points share the same set of coordinates but the fake diagram has random paring.}
\end{figure}

\subsection{Datasets}


We apply permutation test on diverse data types, whose quantitative summaries are in the appendix.

\textbf{Graph Datasets}: We perform experiments on common benchmark datasets for graph classification. \texttt{IMDB-B, IMDB-M, REDDIT 5K} are composed of social networks. \texttt{BZR, COX2, DD, DHFR}, \\\texttt{FRANKENSTEIN, NCI1, PROTEINS, PTC} are graphs from medical or biological frameworks. 

\textbf{Shape Datasets}: We use Princeton Shape Benchmark \cite{chen2009benchmark} for shape segmentation. This benchmark contains 19 categories of different objects (human, cup, glasses...) with 20 shapes for each category. For shape classification, we use ModelNet10/ModelNet40 \cite{wu20153d} where there are 4899/12,311 CAD models from 10/40 man-made object categories (bathtub, bed, chair...) respectively.

\subsection{Choice of Kernels }

We test four kernels for \PDs i.e., sw, pss, pwg and pf for different filtration functions on \texttt{PROTEINS}. As shown in Table \ref{chosekernel}, the performance of different kernels are close in terms of accuracy. However, the computational cost and hyper-parameter search range for different methods are quite different, which can become a bottleneck in our case.



\begin{table}[h]
\centering
\caption{Performance of different filtration function and persistence kernels for \texttt{PROTEINS} dataset.} 
\scalebox{1}{
\begin{tabular}{lrrrrr}
\hline
fil &  cc &  deg & fiedler\_s & ricci &  mean \\
\hline
pf  & 70.3 & 73.4 &    72.8 &  70.6 & 71.78 \\
pss & 74.3 & 72.5 &    73.9 &  73.6 & 73.58 \\
sw  & 74.0 & 73.6 &    73.5 &  73.8 & 73.72 \\
pwg  & 74.7 & 72.4 &    68.4 &  73.8 & 72.32 \\
\hline
\end{tabular}
}
\label{chosekernel}
\end{table}

We prefer Sliced Wasserstein kernel mainly because it is fast ($O(m log m)$ as opposed to $O(m^2)$ for pss), easy to tune (search over 5 values for bandwidth as opposed to 45 hyper-parameter combinations for pwg), and still yields decent performance. Note that in this paper we focus on understanding the extra power persistence pairing brings in, not achieving the state of the art.

\section{Permutation Test for Graphs}
\subsection{Synthetic Graph Data}
In this section, we perform permutation test on synthetic graph dataset where graphs sampled from 2 stochastic block models are classified. In particular, denote $sbm(n_1, n_2, p, q)$ as stochastic block model of two blocks of size $n_1$ and $n_2$, and within each block, the edge probability is $p$, while between blocks, the edge probability is $q$. We sample 1000 graphs for classification from $sbm(100, 50, 0.5, 0.1)$ and $sbm(75, 75, 0.4, 0.2)$. This is a simple classification problem and there are many ways to achieve perfect results.

\begin{figure}[htbp]
\begin{center}
\includegraphics[height=4cm]{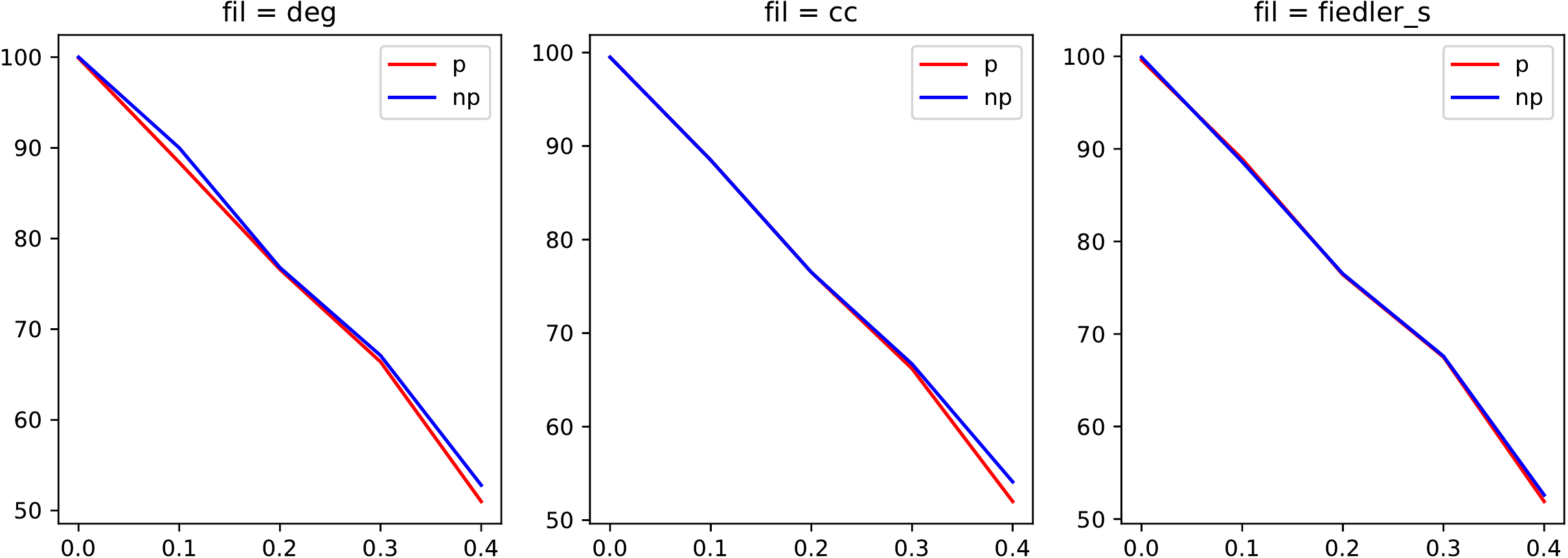} 
\caption{Results on synthetic graph data where the difference between permuting diagram versus not permuting is rather small.}
\label{synfig}
\end{center}
\end{figure}

To make the classification harder, we randomly flip some labels. For example, when label noise is 0.1, we randomly flip 10\% of labels. We are interested in the behaviors of whether to perform permutation test or not under different label noise levels. As shown in Figure \ref{synfig}, independent from  filtration functions used, applying permutation test has rather small effects on the final performance for different noise levels. 

\subsection{Real Graph Datasets} 

In this section, we perform graph classification on common benchmark datasets with different filtration functions and featurizations. Through extensive experiments, we draw the following conclusions.

\textbf{Choice of the Filtration Function.}
The choice of filtration function clearly matters. For datasets such as \texttt{\uppercase{bzr, cox2, dd, dhfr, frankenstein, nci1, reddit 5K}},
 using Ricci curvature as the filtration function yields the best accuracy as opposed to other filtration functions.

If we fix method to be Sliced Wasserstein kernel, for graphs like \texttt{IMDB-B} (69.5 for degree vs. 69.2 for Ricci) and \texttt{IMDB-M} (43.1 vs. 43.7), \texttt{PROTEINS} (73.6 vs. 73.8), \texttt{D\&D} (76.1 vs. 76.9), even degree performs as well as Ricci. This is consistent with the findings in the paper \cite{cai2018simple} where it is shown simple statistics based on node degree can perform on par with the state of the art. This raises the concern that the current benchmark datasets might be limited in evaluating different methods. 

\begin{table}[htp!]
\centering
\caption{The accuracy obtained from filvec, pervec, sw (Sliced Wasserstein Kernel) and sw\_p (with permutation) for different filtration functions and datasets.}
\scalebox{1}{
\begin{tabular}{| c | c | c c c c | c |}
\hline
      graph & fil &  cc &  deg & fiedler\_s & Ricci &  mean \\
\hline
\texttt{\uppercase{bzr}} & filvec & 80.7(0.6) & 83.4(0.5) &    80.5(0.7) &  87.6(0.8) & 83.05 \\
       & pervec & 85.4(0.5) & 82.5(0.7) &    80.5(0.6) &  87.6(0.5) & 84.00 \\
       & sw & 86.2(0.5) & 83.4(0.6) &    81.0(0.4) &  88.4(0.6) & 84.75 \\
       & sw\_p & 85.1(0.5) & 82.2(0.6) &    82.2(0.7) &  82.7(0.7) & 83.05 \\\hline
\texttt{\uppercase{cox2}} & filvec & 78.8(0.5) & 78.2(0.6) &    78.8(0.7) &  82.0(0.6) & 79.45 \\
       & pervec & 78.8(0.6) & 78.2(0.6) &    79.4(0.7) &  81.6(0.5) & 79.50 \\
       & sw & 79.9(0.7) & 78.6(0.5) &    78.2(0.7) &  80.5(0.5) & 79.30 \\
       & sw\_p & 78.2(0.6) & 78.6(0.6) &    78.8(0.5) &  80.1(0.7) & 78.93 \\\hline
\texttt{\uppercase{DD}} & filvec & 75.0(0.4) & 72.5(0.4) &    67.0(0.6) &  76.8(0.6) & 72.82 \\
       & pervec & 75.1(0.4) & 70.5(0.6) &    66.4(0.5) &  77.4(0.5) & 72.35 \\
       & sw & 76.1(0.6) & 76.1(0.5) &    71.1(0.5) &  76.9(0.5) & 75.05 \\
       & sw\_p & 76.7(0.5) & 76.0(0.6) &    74.3(0.4) &  77.4(0.5) & 76.10 \\\hline
\texttt{\uppercase{dhfr}} & filvec & 75.1(0.4) & 67.3(0.6) &    72.1(0.5) &  80.1(0.6) & 73.65 \\
       & pervec & 75.5(0.6) & 71.3(0.5) &    73.3(0.5) &  80.8(0.4) & 75.23 \\
       & sw & 79.0(0.7) & 73.4(0.6) &    74.7(0.4) &  82.8(0.5) & 77.48 \\
       & sw\_p & 78.7(0.5) & 74.7(0.5) &    76.6(0.6) &  76.5(0.6) & 76.62 \\\hline
\texttt{\uppercase{frank}} & filvec & 65.3(0.2) & 66.8(0.3) &    62.8(0.3) &  72.3(0.2) & 66.80 \\
       & pervec & 65.2(0.2) & 65.4(0.2) &    63.0(0.3) &  70.8(0.2) & 66.10 \\
       & sw & 67.8(0.3) & 67.3(0.4) &    67.1(0.3) &  72.0(0.2) & 68.55 \\
       & sw\_p & 65.6(0.2) & 66.8(0.3) &    65.0(0.2) &  69.0(0.3) & 66.60 \\\hline
\texttt{\uppercase{imdb-b}} & filvec & 66.4(0.5) & 64.6(0.6) &    60.4(0.6) &  65.6(0.6) & 64.25 \\
       & pervec & 65.7(0.5) & 67.1(0.6) &    63.1(0.6) &  63.7(0.5) & 64.90 \\
       & sw & 69.5(0.6) & 69.5(0.5) &    66.5(0.6) &  69.2(0.5) & 68.68 \\
       & sw\_p & 66.1(0.5) & 67.8(0.5) &    65.2(0.7) &  67.5(0.5) & 66.65 \\\hline
\texttt{\uppercase{imdb-m}} & filvec & 46.0(0.3) & 46.1(0.3) &    43.5(0.4) &  46.5(0.3) & 45.52 \\
       & pervec & 42.5(0.3) & 42.7(0.3) &    42.1(0.3) &  42.9(0.3) & 42.55 \\
       & sw & 42.6(0.4) & 42.6(0.4) &    45.7(0.2) &  45.2(0.3) & 44.03 \\
       & sw\_p & 42.3(0.3) & 42.3(0.3) &    45.0(0.3) &  42.6(0.2) & 43.05 \\\hline
\texttt{\uppercase{nci1}} & filvec & 69.3(0.2) & 64.7(0.3) &    67.0(0.3) &  74.3(0.3) & 68.82 \\
       & pervec & 69.7(0.2) & 63.4(0.3) &    64.2(0.3) &  74.4(0.1) & 67.93 \\
       & sw & 74.9(0.2) & 67.2(0.2) &    72.1(0.2) &  77.8(0.3) & 73.00 \\
       & sw\_p & 69.9(0.3) & 64.9(0.2) &    69.3(0.2) &  71.1(0.2) & 68.80 \\\hline
\texttt{\uppercase{proteins}} & filvec & 72.4(0.4) & 68.3(0.4) &    71.7(0.3) &  71.2(0.4) & 70.90 \\
       & pervec & 72.6(0.4) & 69.4(0.4) &    71.1(0.4) &  70.9(0.3) & 71.00 \\
       & sw & 74.0(0.4) & 73.6(0.2) &    73.5(0.3) &  73.8(0.3) & 73.72 \\
       & sw\_p & 73.1(0.3) & 72.1(0.3) &    72.8(0.3) &  73.0(0.2) & 72.75 \\ \hline
\texttt{\uppercase{reddit 5K}} & filvec & 47.0(0.1) & 45.8(0.2) &   40.3(0.1) &  51.0(0.2) & 46.03 \\ 
     & pervec & 49.2(0.2) & 48.9(0.1) &   39.0(0.1) &  49.7(0.2) & 46.70 \\
     & sw & 52.6(0.2) & 49.1(0.1) &   42.5(0.1) &  54.1(0.1) & 49.58 \\
     & sw\_p & 50.4(0.1) & 49.2(0.1) &   41.9(0.1) &  53.3(0.2) & 48.70 \\
\hline
\end{tabular}
}
\label{default}
\end{table}%

\begin{table*}[h]
\centering
\caption{Fix featurlization as sw and look at the whether adding $Ext_1^-$ in the \PDs helps. The accuracy takes the maximum over different filtration function.} 
\label{epd}
\scalebox{1}{
\resizebox{1\textwidth}{!}{ 
\begin{tabular}{ l c c c c c c c c c c c c c } \hline
graph &  \uppercase{\texttt{bzr}} & \texttt{\uppercase{cox2}} & \texttt{\uppercase{d\&d}} & \texttt{\uppercase{dhfr}} &  \texttt{\uppercase{frank}} & \texttt{\uppercase{imdb-b}} & \texttt{\uppercase{imdb-m}} &  \texttt{\uppercase{nci1}} & \texttt{\uppercase{protein}} &  \texttt{\uppercase{ptc}} & \texttt{\uppercase{reddit\_5K}} \\ \hline

sw wo/ $Ext_1^-$ &  86.6 &  80.3 &     76.2 &  81.9 &          70.8 &         66.2 &        42.6 &  77.1 &          72.7 &  58.8 &       53.1 \\
sw w/ $Ext_1^-$ &  $\mathbf{88.4}$ &  80.5 &     76.9 &  $\mathbf{82.8}$ &          $\mathbf{72.0}$ &         69.5 &        45.7 &  77.8 &          74.0 &  58.7 &       54.1 \\ \hline

pervec wo/ $Ext_1^-$ &  85.4 &  81.6 &     74.4 &  80.8 &          70.1 &         65.7 &        42.7 &  74.4 &          70.9 &  59.3 &       49.5 \\
pervec w/ $Ext_1^-$  &  87.6 &  $\mathbf{81.6}$ &     77.4 &  80.0 &          70.8 &         67.1 &        42.9 &  74.3 &          72.6 &  59.3 &       49.7 \\ 
\hline
Perslay & 87.2 & $\mathbf{81.6}$ &   - & 81.8 &     70.7 &     70.9 &    48.7 & 72.8 &  74.8 &  - &    56.6 \\ 
WKPI-kM & - & -  &   $\mathbf{82.0}$ & - &- &     70.7 &    46.4 &  $\mathbf{87.5}$   & $\mathbf{78.5}$ &  62.7 &   59.1 \\
WKPI-kC & - & -  &   80.3                    & - &- &   $\mathbf{75.1}$ &    $\mathbf{49.5}$ &    84.5   & 75.2 &  $\mathbf{68.1}$ &   $\mathbf{59.5}$ \\
\hline
\end{tabular}
}
}
\end{table*}

\textbf{Sliced Wasserstein Kernel + Ricci is Powerful.}
Choosing the best accuracy for sw among different filtrations yields decent performance.
\texttt{\uppercase{cox2}}: 80.5 (best accuracy when using sw) vs. 82.0 (best accuracy among all filtration functions and featurlizations for a single dataset).
\texttt{\uppercase{bzr}}: 88.4 vs. 88.4.
\texttt{\uppercase{dd}}: 76.9 vs. 77.4.
\texttt{\uppercase{dhfr}}: 82.8 vs. 82.8.
\texttt{\uppercase{frankenstein}}: 72.0 vs. 72.3.
\texttt{\uppercase{imdb-b}}: 69.2 vs. 69.5.
\texttt{\uppercase{imdb-m}}: 45.2 vs. 46.5.
\texttt{\uppercase{nci1}}: 77.8 vs. 77.8.
\texttt{\uppercase{proteins}}: 73.8 vs. 74.0.
\texttt{\uppercase{reddit 5K}}: 54.1 vs. 54.1.

As a corollary, to achieve good performance for graph classification, a rule of thumb is to use Ricci curvature as filtration plus Sliced Wasserstein kernel.

\textbf{Permutation Test.}
Looking at the mean accuracies over four different filtration functions, sw performs better than sw\_p/pervec/filvec, although the amount of improvement depends on the dataset. 
For \texttt{\uppercase{bzr, dhfr, frankenstein, imdb-b, nci1, protein, reddit 5K}}, sw is better than filvec, pervec, and sw\_p.
For \texttt{\uppercase{cox2, dd}}, sw is no better than the best of filvec/pervec/sw\_p, but the gap (0.2 for \texttt{\uppercase{cox2}} and 1.05 for \texttt{\uppercase{dd}}) is small. 

Pervec and filvec are used as baselines. The performance of pervec, filvec and sw\_p is expected to be close to each other since none of them is using persistence pairing. This is indeed the case. The best of pervec and filvec is close to sw\_p for all the graphs. (The difference is less than 2.5 percent.) Note hyper-parameter choice for pervec/filvec and sw\_p will also result in different performance. After all, the input for pervec/filvec are vectors while for sw\_p the input is fake PDs. We believe that the difference between pervec, filvec and sw\_p is reasonable, and the conclusion that most discriminative power comes from function values is rather robust.


\textbf{Learning for PDs.}
We compare the accuracy obtained from sw + best filtration function with Perslay/WKPI where learning is involved. As shown in Table \ref{epd}, our method is comparable with Perslay. We conjecture replacing original filtration (based on heat kernel signature) used in Perslay with Ricci curvature may improve its performance for some datasets. WKPI (WKPI-kM and WKPI-kC differs in how they initialize weights.) learns the weights of different points in diagrams, which results in much better performance. This confirms the belief that to fully utilize the power of PDs, a proper weighting scheme is crucial.
\ifcomments
\chen{More discussion.} 
\fi


\subsection{The Effect of Adding Loops}
We analyze the effect of adding loops ($Ext_1^-$) in extended \PDs for graph classification. Since extended \PDs capture loops in graphs with respect to the filtration function, the hope is that adding $Ext_1^-$ in the \PD will make it more discriminative. 

But one can perhaps argue the improvement may come from the coordinate values of extended PDs, so we fix featurization as pervec (as opposed to sw) to see the difference. Table \ref{epd} shows that adding coordinates of extended \PDs increase the accuracy, no matter whether we utilize pairing (sw) or not (pervec).

\section{Permutation Test for Shapes}
We now perform permutation test on the problem of supervised 3D shape segmentation and object classification. As we will see, the choice of permuting diagrams makes a much bigger difference here.

\subsection{Shape Segmentation}

For each vertex $x$, we use the geodesic distance (distance to $x$) as the filtration function and compute the super-level 0-\PDs as features. We convert PDs into feature vectors via either persistence landscape \cite{bubenik2015statistical} or persistence image \cite{adams2017persistence} and use SVM \cite{ma2017diving} as our classifier. 

We use Princeton shape benchmark as dataset. This benchmark contains several different ground truth segmentations for each shape. For each shape in the training set, we use the same ground truth segmentation as \cite{kalogerakis2010learning} (that is the segmentation with lowest average Rand Index to all other segmentations for that shape). For each category, we use 50\% data for training and the rest for testing. The results are shown in Table \ref{seg}, from which we draw following conclusions. 

The effects of permutation test clearly depend on the vectorization selected. For Persistence Landscape (PL), there are some categories (such as cup, glasses, vase...) where permuting \PDs yields better results than the original diagrams. We find corresponding diagrams for those categories all have very few persistence points (less than 2) far away from the diagonal on average. See diagram statistics  in the appendix. In the formulation of persistence landscape, points close to diagonal are treated less important and thus play a less important role in final shape segmentation. In contrast, permuting \PDs will ``pull'' those points away from diagonal with high probability and make them more important under PL's framework, therefore explaining the fact that permuting diagrams coupled with persistence landscape yields better results for those categories.
\begin{figure}[htp!]
\centering
\scalebox{1}{
  \begin{subfigure}[t]{.223\textwidth}
    \centering
    \includegraphics[width=\linewidth]{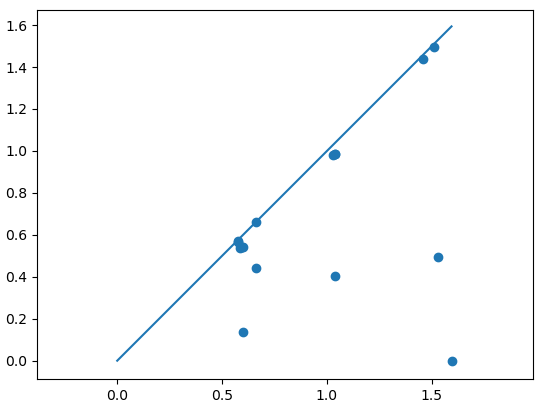}
  \end{subfigure}
  \begin{subfigure}[t]{.223\textwidth}
    \centering
    \includegraphics[width=\linewidth]{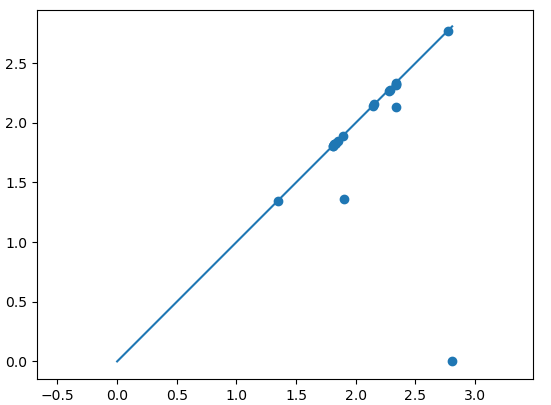}
  \end{subfigure}
  }
  \caption{Two corresponding diagrams for Human and Cup. On average, diagrams in Human/Cup has 4.4/1.5 points away from diagonal.}
\end{figure}
In contrast, Persistence Image (PI) takes a very different way to vectorize \PDs. In particular, PI assigns a weight for each point and points near diagonal are not necessarily assigned small weight. We use the same weight (proportional to death time) as in the original paper.  It turns out that 1) permuting \PDs for PI always gives worse result except for categories bearing and bust where a very small gap exists and 2) PI yields better results on all shapes compared to PL. 
Interestingly, given that PL gives smaller weights to points closer to the diagonal, while PI does not necessarily do (as in this experiments), we think this suggests that while \PD can identify ``features'' of input shapes in a canonical way, the importance of these features (especially in terms of the tasks they are used for) may not be consistent with their persistence. This point has been made earlier in \cite{kusano2016persistence, adams2017persistence, zhao2019learning}, which allow assigning different weights to persistence points. 
In our case, permutation test is proved to be a handy trick that can be applied to quickly test if the downstream featurization is effective. 

\begin{table}
\centering
\caption{The performance (error) of Persistence Landscape (PL) and Persistence Image (PI) for permuting diagrams (P) and not permuting diagrams (NP).} 
\scalebox{1}{
\label{seg}
\begin{tabular} {l|cc|cc}
\hline
 & PL + NP &  PL + P &  PI + NP & PI + P \\ 
 \hline
   Human &  $7.2$ & $20.9$ & $3.2$ & $8.2$ \\
    Cup &  $8.8$ & $5.3$ & 2.6 & 3.6\\
  Glasses & $2.6$ & $2.4$ & 1.6 & 2.0\\
 Airplane &  $9.7$ & $17.6$ & 3.5 & 10.7\\
    Ant &  $2.5$ & $5.0$		& 1.6 & 4.6 \\
   Chair &  $3.7$ & $5.0$ 	& 1.1  & 4.3\\
  Octopus &  $2.8$ & $4.5$ 	& 1.1 & 3.8\\
   Table &  $1.6$ & $1.7$ 	& 0.3 & 0.9\\
   Teddy &  $18.6$ & $16.6$ & 4.0 & 11.3\\
   Hand &  $10.8$ & $19.0$ & 1.8 & 10.3 \\
   Plier &  $3.2$ & $4.9$ 	& 2.9 & 4.4\\
   Fish &  $14.4$ & $10.3$ & 7.7 & 8.1\\
   Bird &  $8.4$ & $10.8$ & 3.0 & 9.4\\
   Armadillo &  $17.1$ & $39.7$ & 4.6 & 23.6\\
    Bust & $44.0$ & $32.5$ & 16.7 & 15.9 \\
    Mech & $23.8$ & $18.3$ & 11.0 & 13.6\\
  Bearing &  $13.6$ & $6.7$ & 3.3  & 2.9\\
    Vase &  $24.3$ & $16.1$ & 7.0 & 9.2\\
  Fourleg &  $13.0$ & $20.7$ & 3.5 & 11.5 \\
\hline
\end{tabular}
}
\end{table}
\subsection{3D Object classification}
Next, we evaluate our method for shape classification on ModelNet10/ModelNet40. For each shape,
 1024 points are uniformly sampled  on mesh faces according to face area and normalized into a unit sphere. We then construct a 8-neighborhood graph from sampled points and use closeness centrality as our filtration function. For each shape, we use the resulting \PDs as the shape representations and use sw/pf/PI/PL for object classification. 

Table \ref{3d} shows that if we do not utilize persistence pairing, we get very similar results for three methods (P, Pervec, Filvec),  which are much worse than result using \PDs (NP), regardless of featurlizations. A detailed performance breakdown on ModelNet 10 is also shown in Table \ref{breakdown} where permuting diagrams yields much worse results on all shape categories except dresser.

\begin{table}[htp!]
\caption{Classification accuracy on ModelNet. Four numbers under NP (no permutation) and P (permutation) are accuracies for sw, pf, PI and PL.}
\centering
\label{3d}
\begin{tabular}{lllll}
\hline
 &  NP &  P & Pervec & Filvec \\ \hline
   ModelNet-10 &  $55.2/53.6/53.1/53.2$ & $42.7/42.9/41.5/41.5$ & 43.5 & 44.5 \\ 
   ModelNet-40 &  $43.4/35.8/35.7/34.2$ & $28.1/27.5/27.1/26.9$ & 28.5 & 30.1 \\ \hline
\end{tabular}
\end{table}
\ifcomments
\chen{todo. Insights.}
\fi
Results both in Table \ref{seg} and \ref{3d}   suggest that persistence pairings are  effective and meaningful for  3D models, which partly could be due to that 3D models tend to have clear geometric features that are also stable under the typical types of (Hausdorff) noise added to these models. 

\begin{table*}[h]
\caption{The performance ($f1$ score) breakdown of using \PDs for 3D object classification on ModelNet10.}
\centering
\scalebox{.8}{
\label{breakdown}
\begin{tabular}{@{}llllllllllll@{}}
\hline
          & mean         & bathtub & bed & chair & desk & dresser & monitor & night stand & sofa & table & toilet \\ \hline
\# of shapes     & - & 156   & 615 & 989  & 286 & 286   & 565   & 286     & 780 & 492  & 444  \\
$f1$ score wo/ permutation & 57.2          & 54.1   & 56.4 & 76.7  & 21.2  & 36.3   & 64.5   & 46.8     & 62.1  & 54.2  & 74.0   \\
$f1$ score w/ permutation & 44.3          & 24.3   & 37.3 & 63.7  & 0.0  & 46.1   & 44.7   & 31.2     & 48.8  & 16.1  & 15.2   \\ \hline
\end{tabular}
}
\end{table*}

In contrast, persistent homology seems to be less effective at capturing features for graphs: this could partly be due to the choice of descriptor functions used for graphs are not effective at capturing features. Another potential reason could be that \PDs are more sensitive to the common types of noise in graphs (random insertions), and hence resulting persistence-based features are less stable and meaningful.

\section{Analysis and Visualization}
\subsection{Separation from True and Fake Diagrams}
Due to the decent result without using persistence pairing on graph datasets, one may wonder whether there is any useful structure contained in persistence pairing for graph classification. To better understand the implications of the permutation test, we try to separate true PDs from fake PDs. 
\begin{table}[htp!]
\caption{The accuracy of seprating true PDs from fake ones for different graphs and filtration functions. } 
\centering
 \scalebox{1}{
\begin{tabular}{ c c c c } 
\hline 
& deg & ricci & cc  \\ \hline 
\texttt{\uppercase{Imdb-b}}  & 99.8 & 99.5 & 99.6 \\ 
\texttt{\uppercase{Imdb-m}}  & 99.8 &  99.8  & 99.7   \\ 
\texttt{\uppercase{reddit-b}} & 95.7 & 87.6   & 91.7   \\ 
\texttt{\uppercase{dd}}      & 99.5 & 99.3 & 99.6  \\ \hline 
\end{tabular}
\label{sepration}
}
\end{table}
For each dataset and filtration function, we compute \PDs and generate fake diagrams by applying permutation test. We compute the Sliced Wasserstein kernel and train SVM to discriminate the true diagrams from the fake ones. We observe in Table \ref{sepration} consistently that regardless of the datasets and filtration functions, we can easily get 85\%-100\% accuracy. This suggests that true diagrams have some structure that is very different from fake ones.

\subsection{Confusion Matrix Analysis}
We also examine the confusion matrix in cases where we need to additionally differentiate true/fake diagrams. In particular, for each graph $G_{i}$ with label $y_{i}$ (assume all labels are represented as positive numbers), we compute the true \PD of each graph and generate a fake diagram. In scenario 1) we assign both diagrams as same label $y_{i}$ and in scenario 2), we assign two diagram with different label $y_{i}$ and $-y_{i}$. In other words, in the second scenario, a classifier has to differentiate both true/fake diagrams and diagrams of different types of graphs. We want to know whether scenario 2 will make problem harder.

\begin{table}[h]
\caption{Confusion matrices obtained from kernel SVM without (left)/with (right) fake \PDs on \texttt{DD}. I/II stands for graph types. T/F stands for whether \PDs used are true or fake.}\smallskip
\centering
\begin{tabular}{cccccccc}
 & I & II &  & I+T & I+F & I+F & II+F \\ \cline{2-3} \cline{5-8} 
\multicolumn{1}{c|}{I} & \multicolumn{1}{c|}{121} & \multicolumn{1}{c|}{22} & \multicolumn{1}{c|}{I+T} & 65 & \multicolumn{1}{c|}{0} & 8 & \multicolumn{1}{c|}{0} \\ \cline{2-3}
\multicolumn{1}{c|}{II} & \multicolumn{1}{c|}{38} & \multicolumn{1}{c|}{55} & \multicolumn{1}{c|}{I+F} & 0 & \multicolumn{1}{c|}{56} & 0 & \multicolumn{1}{c|}{14} \\ \cline{2-3} \cline{5-8} 
 &  &  & \multicolumn{1}{c|}{II+T} & 25 & \multicolumn{1}{c|}{0} & 29 & \multicolumn{1}{c|}{0} \\
 &  &  & \multicolumn{1}{c|}{II+F} & 0 & \multicolumn{1}{c|}{13} & 0 & \multicolumn{1}{c|}{26} \\ \cline{5-8} 
\end{tabular}
\label{confusion}
\end{table}

In particular, we use node degree as filtration function and Sliced Wasserstein kernel. 
It can be seen in Table \ref{confusion} that fake diagrams never get confused with true diagrams. We can recover the confusion matrix on the left from the matrix on the right by merging true and fake diagrams.


\begin{figure}[h]
\centering
\includegraphics[width=.8\textwidth]{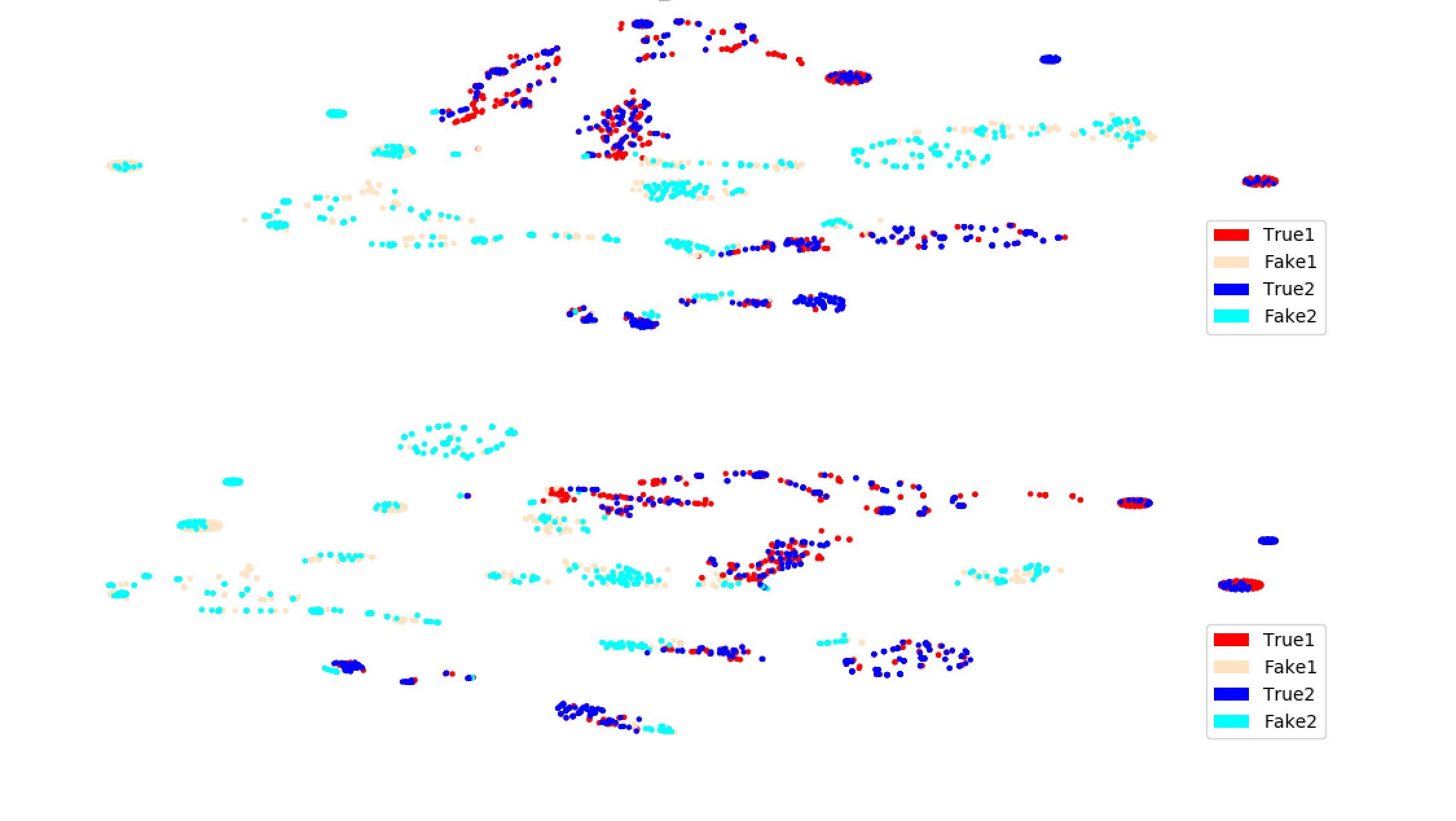} 
\caption{The visualization of true/fake \PDs in dark/bright color for \texttt{IMDB-B}. Top: TSNE with kernel distance induced from Sliced Wasserstein kernel. Bottom: TSNE with bottleneck distance (see appendix).}
\label{fig2}
\end{figure}

\section{Conclusion and Future Work}

By introducing permutation test on \PDs, we find persistence pairing is crucial for shape datasets. For graph datasets, although the small (yet consistent) improvements pairing brings may be unexpected and unsatisfying, we interpret this due to the challenging nature of graph classification and dataset problem. 


In the future, we are interested in understanding the discriminative power of \PDs in a principle way. For example, can we make reasonable generative models of graphs/shapes under which we can theoretically quantify the improvement of persistence paring? We believe developing a connection between the structure of persistence module (persistence pairing) and generalization error is important for the application of \PDs in machine learning.

\bibliography{ref}

\newpage
\appendix
\section{Concepts}
Due to the space limitation, we list definitions of some concepts needed for the paper in the appendix. We refer readers to \cite{edelsbrunner2010computational, oudot2015persistence} for more details.
\subsection{Homology}
The key concept of homology theory is to study the properties of some object $X$ by means of (commutative) algebra. In particular, we 
assign to $X$ a sequence of modules $C_0, C_1, ...$ which are connected 
by homomorphisms 
$\partial_{n}:C_{n} \rightarrow C_{n-1}$ 
such that $\text{im} \partial_n \subseteq \text{ker} \partial_n$
A structure of this form is called a chain complex and by studying its homology groups 
$H_{n}=\operatorname{ker} \partial_{n} / \operatorname{im} \partial_{n+1}$
we can derive properties of $X$.


\subsection{Bottleneck Distance}
Given two PDs $D_1$ and $D_2$, let $\Gamma : D_1 \supseteq A \rightarrow B \subseteq D_2$ be a partial bijection between $D_1$ and $D_2$. Then for any point $x\in A$, the \emph{p-cost} of $x$ is defined as $c_{p}(x)=\|x-\Gamma(x)\|_{\infty}^{p}$ and 
for $y \in (D_1 \cup D_2 ) \backslash (A \cup B) $, the $p$-cost of $y$ is defined as $c_{p}^{\prime}(y)=\left\|y-\pi_{\Delta}(y)\right\|_{\infty}^{p}$, where $\pi_{\Delta}$ the projection onto the diagonal $\Delta =\{(x, x) : x \in \mathbb{R}\}$. The cost of this partial bijection $\Gamma$ is defined as $c_{p}(\Gamma)=\left(\sum_{x} c_{p}(x)+\sum_{y} c_{p}^{\prime}(y)\right)^{1 / p}$. Finally, define the $p$-th diagram distance $d_p$ as the cost of the best partial bijection:
\begin{equation}
d_{p}\left(D_{1}, D_2\right)=\inf _{\Gamma} \mathrm{c}_{p}(\Gamma), 
\end{equation}
where $\Gamma$ ranges over all partial bijections between $D_1$ and $D_2$. 
In the particular case when $p = \infty$, the cost of $\Gamma$ is therefore $ c(\Gamma)=\max\{\max _{x} c_{1}(x)+\max _{u} c_{1}^{\prime}(y)\} )$. The corresponding distance $d_{\infty}$ is often called the bottleneck distance between diagrams $D_1$ and $D_2$. 

\section{Diagram Statistics}
We list diagram statistics for the shape segmentation task in Table \ref{stat}. Ave \# of PD points (first row) stands for the average number of persistence points in the diagram for a shape category. A persistence point is considered as near diagonal if its lifetime is less than one-tenth of the lifetime of the furthest point in the same diagram. 

As we can see in Table \ref{stat}, most persistence points for mech, bust, cup, bearing, glasses, fish, vase, teddy are concentrated near diagonal. Those are exactly the same categories on which permuting diagram yields much better results than not permuting diagrams when PI is used.

\begin{table*}[htp]
\caption{The diagram statistics for different shape categories in Princeton Shape Benchmark. } \smallskip
\scalebox{0.55}{
\begin{tabular}{llllllllllllllllllll}
\hline
Categroy    & Mech & Bust &  Cup & Bearing & Glasses & Fish & Vase & Teddy & Bird & Plier & Airplane & Table & Human & Armadillo & Chair & Hand & Fourleg & Octopus &  Ant \\
\hline
Ave \# of PD points   &  11 & 22.8 & 15.9 &   16.9 &   7.5 & 15.1 & 15.7 &  18.3 &  18 &  6.6 &   12.5 &  16.6 &  58.2 &    78.6 &  18.8 & 19.7 &   31.4 &   14.6 & 15.9 \\
\# of points near diagonal&  9.9 & 21.5 & 14.4 &   15.3 &   5.6 & 13.2 & 13.7 &  15.1 & 14.3 &  2.9 &    8.5 &  12.3 &  53.7 &    74.1 &  14.1 & 14.5 &   25.9 &   6.7 &  7.3 \\
Difference   &  1.1 &  1.3 &  1.5 &   1.6 &   1.9 &  1.9 &   2 &  3.2 &  3.6 &  3.7 &     4 &  4.3 &  4.4 &    4.6 &  4.7 &  5.1 &   5.4 &   7.9 &  8.6 \\
\hline
\end{tabular}
}
\label{stat}
\end{table*}%

\section{Vector Methods for \PDs}
\textbf{Persistence Landscape} \cite{bubenik2015statistical} is the first proposed vectorization method for \PDs to overcome some undesirable property of the space of \PDs, such as lacking a unique Frechet mean. This construction is mainly intended for statistical computations, enabled by the vector space structure of $L_{p}$. Given a \PD $D = \{(b_{i}, d_{i})\}_{i=1}^{m}$, persistence landscape can be thought of as a sequence of functions $\lambda_{k}: \mathbb{R} \rightarrow \mathbb{\bar{R}}$ where $\lambda_{k}(t) =$ $k$th largest value of min $(t-b_{i}, d_{i}-t)_{+}$.

\textbf{Persistent Image} \cite{adams2017persistence} produces a persistence surface $\rho_{B}$ from a \PD by taking a weighted sum of Gaussians centered at each point. The vector representation, named by persistence image, is created by integrating persistence surface over a grid. In particular, they fix a grid in the plane with $n$ pixels and assign to each the integral of $\rho_{B}$ over that region.

There are at least three parameters involved in the construction of persistence image: 1) a non-negative weighting function 2) the bandwidth of Gaussian kernel (many other functions can be chosen but in the original paper only Gaussian is considered) and 3) the resolution of the grid put over the persistence surface. The authors report that in classification experiments they conducted, the accuracy is insensitive to the choice of resolution and bandwidth.


\section{Kernels Methods for PDs}
\textbf{Persistence Weighted Gaussian Kernel} \cite{kusano2016persistence}
essentially utilizes the idea of kernel mean embedding of distribution, where persistence diagram, treated as a special case of distribution, can be embedded into RKHS. In particular, Let $K, \rho > 0$ and $D_{1}$ and $D_{2}$ be two PDs. Let $K_{\rho}$ be the Gaussian kernel with parameter $\rho>0 $. Let $H_{\rho}$ be the RKHS associated to $k_{\rho}$ .

Let $\mu_{1} = \Sigma _{x \in D_{1}} \text{arctan}(K \textit{pers}(x)^{p} k_{\rho}(*, x)) \in H_{p}$ be the kernel mean embedding of $D_{1}$ weighted by the diagonal distances. Let $\mu_{2}$ be defined similarly. Let $\tau>0$, the persistence weighted gaussian kernel $K_{pwg}$ is defined as the gaussian kernel with bandwidth $\tau$ on $H_{p}:$
\begin{equation}
K_{pwg} (D_{1}, D_{2}) = e^{- \frac{\| \mu_{1} - \mu_{2}\|_{H_{p}}}{2t^{2}}}
\end{equation}

\textbf{Persistence Scale Space Kernel} \cite{reininghaus2015stable} represents persistence diagram as sum of Dirac's delta measure. The persistence scale space kernel is defined as the scalar product of the solution of the heat diffusion equation with the persistence diagram as an initial value.

The closed form 
\begin{equation}
 K_{pss}(D_{1}, D_{2}) = \frac{1}{8} \sum\limits_{p \in D_{1}} \sum\limits_{q \in D_{2}} e^{(-\frac{\|p-q\|^{2}}{8t})}-e^{(-\frac{\|p-\bar{q}\|^{2}}{8t})}
\end{equation}
can be computed exactly in $O(|D_1|*|D_2|)$ time where $\bar{q}=(y, x)$ is the symmetric of $q=(x, y)$ along the diagonal. $|D_1|$ and $|D_2|$ denote the cardinality of the multisets $D_{1}$ and $D_{2}$

\textbf{Sliced Wasserstein Kernel} \cite{carriere2017sliced} uses Sliced Wasserstein approximation of the Wasserstein distance to define a new kernel for PDs. Different from previous multiple kernels, it is provable not only \emph{stable} but also \emph{discriminative} (with a bound depending on the number of points in the PDs) w.r.t. the first diagram distance $w_1^{\infty}$ between PDs. 

In particular, the kernel has the following closed-form:
\begin{equation}
K_{sw} (D_{1}, D_{2}) = e^{\frac{-SW(D_{1}, D_{2})}{2\sigma^{2}}}
\end{equation}
where $SW(D_{1}, D_{2})$, is defined as the sliced Wasserstein distance between PDs. 

\textbf{Persistence Fisher Kernel} \cite{le2018persistence} differs from slice Wasserstein kernel in the sense that the Wasserstein geometry is replaced by Fisher information geometry (metric), which induces a negative definite distance. The form of Persistence fisher kernel is
\begin{equation}
k_{PF}(D_1, D_2) = e^{-t d_{FIM}(D_1, D_2)}
\end{equation}
where $t$ is a positive scalar and $d_{FIM}$ is the Fisher information metric.

\textbf{Weighted Persistence Image Kernel} (WKPI) \cite{zhao2019learning} is recently proposed weighted kernel that is based on persistence image. In particular, WKPI assigns weight for different location in the diagram where the weight is learned via gradient descent. The form of WKPI is
\begin{equation}
k_{w}(PI, PI') = \Sigma_{s=1}^{N} w(p_s)e^{-\frac{(PI(s) - PI'(s) )^2}{2 \sigma^2}}
\end{equation}
where PI, PI' are persistence image, $s$ is the location for each pixel in PI, $w$ is the weight function that will be learned from data. Note that majority of time to compute WKPI is spent on learning $w(p_s)$ via stochastic gradient descent. 

\begin{table}[h]
\centering
\caption{Summary of different kernels for PDs. Here $n$ is the number of diagrams. $m$ is the number of persistence points in the diagram of largest size. $M_1$ is the number of random Fourier feature used for approximating Gaussian kernel. $M_2$ is the number of directions for approximating Sliced Wasserstein kernel.} \smallskip %
\begin{tabular}{@{}lllllll@{}}
\hline
                     & pss    & pwg    & sw   & pf & wkpi\\ \hline
\# of Param           & 1     & 3     & 1     & 2 & 2\\
Exact Comp. Time & $O(m^{2}n^{2})$    & $O(m^{2}n^{2})$ & $O(m^2log(m)n^{2})$  & $O(m^2n^2)$ & -\\
Approx. Time & -     & $O(M_{1}mn + M_{1}n^2)$    & $O(M_{2}mlog(m)n^{2})$  & $O(mn^2) $	& -\\ \hline
\end{tabular}
\end{table}

\section{Hyper-parameters Choice}
We chose our hyper-parameters by 10-fold cross-validation on training set. We list the specific search range for all methods below.

\emph{Persistence Landscape:} the number of $\lambda_{k}(t)$ are chosen from \{3, 4, 5, 6, 7, 8\} and each function $\lambda_{k}(t)$ is discretized into \{50, 100, 200, 300\} bins. 

\emph{Persistence Image:} we use the same weight function (Gaussian) in the original paper, the resolution chosen from \{20*20, 30*30\}, and bandwidth of Gaussian is selected from \{0.01, 0.1, 1, 10, 100\}.  

\emph{Persistence Scale Space Kernel:} the parameter $t$ is chosen from 13 values: \{0.001, 0.005, 0.01, 0.05, 0.1, 0.5, 1, 5, 10, 100, 500, 1000\}.

\emph{Sliced Wasserstein Kernel:} following the original paper \cite{carriere2017sliced}, we grid search bandwidth from the 5 values: \{0.01, 0.1, 1, 10, 100\} and the number of slices, i.e., $M_2$ is set to be 10. 

\emph{Persistence Weighted Gaussian Kernel:} we try 
all the combinations of 5 values from \{0.01, 0.1, 1, 10, 100\} for bandwidth, 
and 3 values from \{0.1, 1, 10\} for both $K$ and $\rho$, leading to 45 different sets of parameters.

\emph{Persistence Fisher Kernel:} there are two hyper-parameters $t$ and $\tau$ (for smoothing persistence diagram), both of which are selected from \{0.01, 0.05, 0.1, 0.5, 1, 5, 10, 50, 100\}.

\emph{Pervec/Filvec}: The length of the vector is chosen from \{100, 200, 300\}.

\emph{Choice of SVM hyper-parameter:}
For kernel SVM, the only hyper-parameter $C$ is selected from \{0.01, 1, 10, 100, 1000\}. For Pervec and Filvec, Gaussian kernel is utilized where the bandwidth is selected from \{0.01, 0.1, 1, 10, 100\}. 

\emph{Graph Classification:}
We follow the standard protocol in graph classification literature, i.e., 10-fold cross-validations, using 9 folds for training and the rest for testing, and repeat the experiments 10 times. We report the average classification accuracies. 

\section{Datasets Description}
\label{sec: dataset}
The statistics of the benchmark graph datasets used in the paper are reported in Table 6. We describe these datasets in detail in the next section.
\subsection{Non-attributed Graph Datasets}
\textbf{IMDB-BINARY} \cite{yanardag2015deep} is a movie collaboration dataset that consists of the ego-networks of 1,000 actors/actresses who played roles in movies in IMDB. In each graph, nodes represent actors/actresses, and there is an edge between them if they appear in the same movie. These graphs are derived from
the Action and Romance genres.

\textbf{IMDB-MULTI} is generated in a similar way to IMDB-BINARY. The difference is that it is derived from three genres: Comedy, Romance, and Sci-Fi.

\textbf{REDDIT-BINARY} consists of graphs corresponding to online discussions on Reddit. In each graph, nodes represent users, and there is an edge between them if at least one of them responds to the other's comment. There are four popular subreddits, namely, IAmA, AskReddit, TrollXChromosomes, and atheism. IAmA and AskReddit are two question/answer based subreddits, and TrollXChromosomes and atheism are two discussion-based subreddits. A graph is labeled according to whether it belongs to a question/answer-based community or a discussion-based community.

\textbf{REDDIT-MULTI(5K)} is generated in a similar way to REDDIT-BINARY. The difference is that there are five subreddits involved, namely, worldnews, videos, AdviceAnimals, aww, and mildlyinteresting. Graphs are labeled with their corresponding subreddits.


\subsection{Attributed Graphs}


\textbf{PTC} \cite{helma2001predictive} consists of graph representations of chemical molecules. In each graph, nodes represent atoms, and edges represent chemical bonds. Graphs are labeled according to carcinogenicity on rodents, divided into male mice (MM), male rats (MR), female mice (FM), and female rats (FR).

\textbf{PROTEINS} \cite{borgwardt2005protein} consist of graph representations of proteins. Nodes represent secondary structure elements (SSE), and there is an edge if they are neighbors along the amino acid sequence or one of three nearest neighbors in space. The discrete attributes are SSE types. The continuous attributes are the 3D length of the SSE. Graphs are labeled according to which EC top-level class they belong to.

\begin{table}[htp!]
\centering
\caption{Statistics of the benchmark graph datasets} \smallskip
\scalebox{1}{
\begin{tabular}{@{}llllll@{}}
\hline
Datasets   & graph \# & class \# & average\_nodes \# & average edges \# & label \# \\ \hline

BZR			&405		&2	&35.75	&38.36 &+ \\
COX2 		&467 	&2 	&41.22 	&43.45 &+ \\
DD      & 1178   & 2    & 284.32      & 715.66      & +    \\
DHFR		&467 	&2	&42.43	&44.54 &+ \\
FRANKSTEIN & 4337 & 2 & 16.90	&17.88 & - \\
IMDB BINARY  & 1000   & 2    & 19.77       & 96.53      & -    \\
IMDB MULTI  & 1500   & 3    & 13.00       & 65.94      & -    \\
NCI1     & 4110   & 2    & 29.87       & 32.30      & +    \\ 
PROTEINS    & 1113   & 2    & 39.06       & 72.82      & +    \\
PTC      & 344   & 2    & 14.29       & 14.69      & +    \\
REDDIT 5K   & 4999   & 5    & 508.82      & 594.87      & -    \\

\hline
\end{tabular}
}
\end{table}
\textbf{DD} \cite{dobson2003distinguishing} consists of graph representations of 1,178 proteins. In each graph, nodes represent amino acids, and there is an edge if they are less than six Angstroms apart. Graphs are labeled according to whether they are enzymes or not.

\textbf{NCI1} \cite{shervashidze2011weisfeiler, kriege2012subgraph} consists of graph representations of 4,110 chemical compounds screened for activity against non-small cell lung cancer and ovarian cancer cell lines, respectively.

\textbf{FRANK} \cite{kazius2005derivation} is a chemical molecule dataset that consists of 2,401 mutagens and 1,936 nonmutagens. Originally, nodes are associated with chemical atom symbols.

\textbf{BZR}, \textbf{COX2}, and \textbf{DHFR} \cite{sutherland2003spline}  all are chemical compound datasets. Still, in each graph, nodes
represent atoms, and edges represent chemical bonds. The discrete attributes correspond to atom
types. The continuous attributes are 3D coordinates.

\end{document}